# Intelligent Systems: Architectures and Perspectives

**Ajith Abraham**

Faculty of Information Technology, School of Business Systems
Monash University (Clayton Campus), Victoria 3168, Australia
Email: ajith.abraham@ieee.org, URL: http://ajith.softcomputing.net

**Abstract:** The integration of different learning and adaptation techniques to overcome individual limitations and to achieve synergetic effects through the hybridization or fusion of these techniques has, in recent years, contributed to a large number of new intelligent system designs. Computational intelligence is an innovative framework for constructing intelligent hybrid architectures involving Neural Networks (NN), Fuzzy Inference Systems (FIS), Probabilistic Reasoning (PR) and derivative free optimization techniques such as Evolutionary Computation (EC). Most of these hybridization approaches, however, follow an ad hoc design methodology, justified by success in certain application domains. Due to the lack of a common framework it often remains difficult to compare the various hybrid systems conceptually and to evaluate their performance comparatively. This chapter introduces the different generic architectures for integrating intelligent systems. The designing aspects and perspectives of different hybrid archirectures like NN-FIS, EC-FIS, EC-NN, FIS-PR and NN-FIS-EC systems are presented. Some conclusions are also provided towards the end.

## 1. Introduction

In recent years, several adaptive hybrid soft computing [108] frameworks have been developed for model expertise, decision support, image and video segmentation techniques, process control, mechatronics, robotics and complicated automation tasks. Many of these approaches use a combination of different knowledge representation schemes, decision making models and learning strategies to solve a computational task. This integration aims at overcoming the limitations of individual techniques through hybridization or the fusion of various techniques. These ideas have led to the emergence of several different kinds of intelligent system architectures [14][51-53][58][66][69][92].

It is well known that intelligent systems, which can provide human-like expertise such as domain knowledge, uncertain reasoning, and adaptation to a noisy and time-varying environment, are important in tackling practical computing problems. In contrast with conventional artificial intelligence techniques which only deal with precision, certainty and rigor, the guiding principle of soft computing is to exploit the tolerance for imprecision, uncertainty, low solution cost, robustness, partial truth to achieve tractability, and better rapport with reality [108]. In general hybrid soft computing consists of 4 essential paradigms: NN, FIS, EC and PR. Nevertheless, developing intelligent systems by hybridization is an open-ended rather than a conservative concept. That is, it is evolving those relevant techniques together with the important advances in other new computing methods [35][96]. Table 1 lists the three principal ingredients together with their advantages [12][42].

Table 1. Comparison of different intelligent systems with classical approaches[†]

|  | FIS | NN | EC | Symbolic AI |
|---|---|---|---|---|
| Mathematical model | SG | B | B | SB |
| Learning ability | B | G | SG | B |
| Knowledge representation | G | B | SB | G |
| Expert knowledge | G | B | B | G |
| Nonlinearity | G | G | G | SB |
| Optimization ability | B | SG | G | B |
| Fault tolerance | G | G | G | B |
| Uncertainty tolerance | G | G | G | B |
| Real time operation | G | SG | SB | B |

[†]Fuzzy terms used for grading are good (G), slightly good (SG), slightly bad (SB) and bad (B).

To achieve a highly intelligent system, a synthesis of various techniques is required. Figure 1 shows the synthesis of NN, FIS and EC and their mutual interactions leading to different architectures. Each technique plays a very important role in the development of different hybrid soft computing architectures. Experience has shown that it is crucial, in the design of hybrid systems, to focus primarily on the integration and interaction of different techniques rather than to merge different methods to create ever-new techniques. Techniques already well understood should be applied to solve specific domain problems within the system. Their weaknesses must be addressed by combining them with complementary methods.

Neural networks offer a highly structured architecture with learning and generalization capabilities, which attempts to mimic the neurological mechanisms of the brain. NN stores knowledge in a distributive manner within its weights which have been determined by learning from known samples. The generalization ability of new inputs is then based on the inherent algebraic structure of the NN. However it is very hard to incorporate human *a priori* knowledge into a NN mainly because the connectionist paradigm gains most of its strength from a distributed knowledge representation.

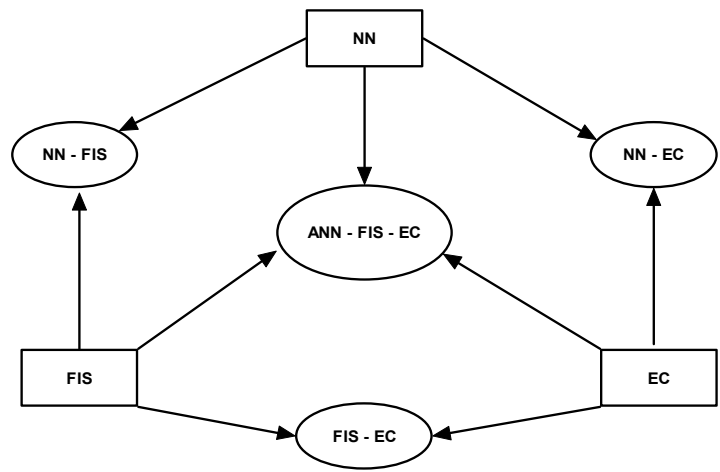

**Figure 1.** General framework for hybrid soft computing architectures

By contrast, fuzzy inference systems [106-107] exhibit complementary characteristics, offering a very powerful framework for approximate reasoning which attempts to model the human reasoning process at a cognitive level [61]. FIS acquires knowledge from domain experts which is encoded within the algorithm in terms of the set of *if-then* rules. FIS employ this rule-based approach and interpolative reasoning to respond to new inputs [30]. The incorporation and interpretation of knowledge is straightforward, whereas learning and adaptation constitute major problems.

Probabilistic reasoning such as Bayesian belief networks [20] and the Dempster-Shafer theory of belief [36] [86], gives us a mechanism for evaluating the outcome of systems affected by randomness or other types of probabilistic uncertainty. An important advantage of probabilistic reasoning is its ability to update previous outcome estimates by conditioning them with newly available evidence [57].

Global optimization involves finding the absolutely best set of parameters to optimize an objective function. In general, it may be possible to have solutions that are locally but not globally optimal. Consequently, global optimization problems are typically quite difficult to solve exactly: in the context of combinatorial problems, they are often NP-hard. Evolutionary Computation works by simulating evolution on a computer by iterative generation and alteration processes operating on a set of candidate solutions that form a population. The entire population evolves towards better candidate solutions via the selection operation and genetic operators such as crossover and mutation. The selection operator decides which candidate solutions move on into the next generation and thus limits the search space [40].

Section 2 presents the various techniques to forumlate hybrid intelligent architectures followed by optimization of neural network using evolutionary computation and local search techniques in Section 3. Adaptation issues of fuzzy inference systems are discussed in Section 4 followed by evolutionary fuzzy systems and cooperative neuro-fuzzy systems in Section 5 and 6 respectively. Integrated neuro-fuzzy systems are presented in Section 7. In Section 8, a framework for an integrated neuro-fuzzy-evolutionary system is presented. Optimization of evolutionary algorithms using soft computing techniques is presented in Section 9 and finally interactions between soft computing technology and probabilistic reasoning techniques are given in Section 10. Some conclusions are also presented.

## 2. Models Of Hybrid Soft Computing Architectures

We broadly classify the various hybrid intelligent architectures into 4 different categories based on the system's overall architecture: (1) Stand-alone (2) Transformational (3) Hierarchical hybrid and (4) Integrated hybrid. The following sections discuss each of these strategies, the expected uses of the model and some benefits and limitations of the approach.

### 2.1 Stand Alone Intelligent System

Stand-alone models consist of independent software components which do not interact in any way. Developing stand-alone systems can have several purposes: first, they provide a direct means of comparing the problem solving capabilities of different techniques with reference to a certain application [13]. Running different techniques in a parallel environment permits a loose approximation of integration. Stand-alone models are often used to develop a quick initial prototype, while a more time-consuming application is developed. Figure 2 displays a stand-alone system where a neural network and a fuzzy system are used separately.

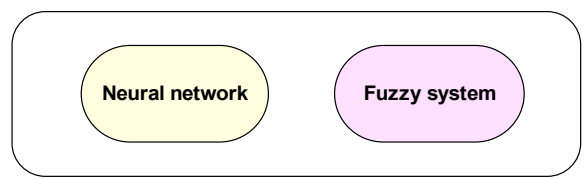

**Figure 2**. Stand–alone system

Some of the benefits are simplicity and ease of development by using commercially available software packages . On the other hand, stand-alone techniques are not transferable: neither can support the weakness of the other technique.

## 2.2 Transformational Hybrid Intelligent System

In a transformational hybrid model, the system begins as one type and ends up as the other. Determining which technique is used for development and which is used for delivery is based on the desirable features that the technique offers. Figure 3 shows the interaction between a neural network and an expert system in a transformational hybrid model [69]. Obviously, either the expert system is incapable of adequately solving the problem, or the speed, adaptability, and robustness of neural network is required. Knowledge from the expert system is used to determine the initial conditions and the training set for the artificial neural network.

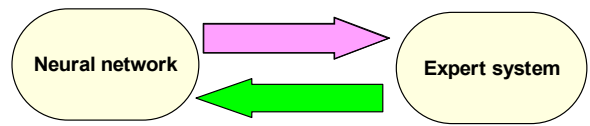

**Figure 3.** Transformational hybrid architecture

Transformational hybrid models are often quick to develop and ultimately require maintenance on only one system. They can be developed to suit the environment and offer many operational benefits. Unfortunately, transformational models are significantly limited: most are just application-oriented. For a different application, a totally new development effort might be required such as a fully automated means of transforming an expert system to a neural network and vice versa.

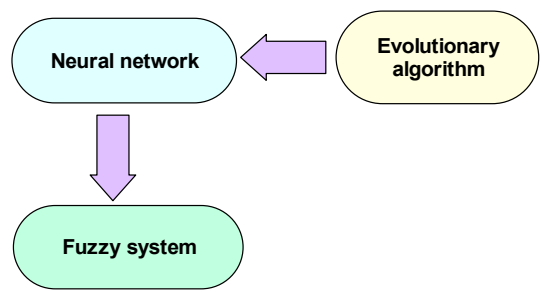

**Figure 4.** Hierarchical hybrid architectures

## 2.3 Hierarchical Hybrid Intelligent System

This architecture is built in a hierarchical fashion, associating a different functionality with each layer. The overall functioning of the model depends on the correct functioning of all the layers. Figure 4 demonstrates a hierarchical hybrid architecture involving a neural network, an evolutionary algorithm and a fuzzy system. The neural network uses an evolutionary algorithm to optimize its performance and the network output acts as a pre-processor to a fuzzy system, which then produces the final output. Poor performance in one of the layers directly affects the final output.

## 2.4 Integrated Intelligent System

Fused architectures are the first true form of integrated intelligent systems. They include systems which combine different techniques into one single computational model. They share data structures and knowledge representations. Another approach is to put the various techniques side-by-side and focus on their interaction in a problem-solving task. This method can allow for integrating alternative techniques and exploiting their mutuality. Furthermore, the conceptual view of the agent allows one to abstract from the individual techniques and focus on the global system behavior, as well as to study the individual contribution of each component [51].

The benefits of integrated models include robustness, improved performance and increased problem-solving capabilities. Finally, fully integrated models can provide a full range of capabilities such as adaptation, generalization, noise tolerance and justification. Fused systems have limitations caused by the increased complexity of the inter-module interactions and specifying, designing, and building fully integrated models is complex. In this chapter, discussions is limited to different integrated intelligent systems involving neural networks, fuzzy inference systems, evolutionary algorithms and probabilistic reasoning techniques.

## 3. Neural Networks and Evolutionary Algorithms

Even though artificial neural networks are capable of performing a wide variety of tasks, in practice, they sometimes deliver only marginal performance. Inappropriate topology selection and learning algorithms are frequently blamed. There is little reason to expect to find a uniformly best algorithm for selecting the weights in a feedforward artificial neural network [97]. It is an NP-complete problem to find a set of weights for a given neural network and a set of training examples to classify even two-thirds of them correctly. In general, claims in the literature on training algorithms that one being proposed is substantially better than most others should be treated with scepticism. Such claims are often defended through simulations based on applications in which the proposed algorithm performed better than some familiar alternative.

The artificial neural network (ANN) methodology enables the design of useful nonlinear systems accepting large numbers of inputs, with the design based solely on instances of input-output relationships. For a training set $T$, consisting of $n$ argument value pairs, and given a $d$-dimensional argument $x$, an associated target value $t$ will be approximated by the neural network output. The function approximation could be represented as:

$$T = \{(x_i, t_i), i = 1 : n\} \tag{1}$$

In most applications, the training set *T* is considered to be noisy and while the goal is not to reproduce it exactly the intention is to construct a network function that generalizes well to new function values. An attempt will be made to address the problem of selecting the weights to learn the training set. The notion of closeness on the training set *T* is typically formalized through an error function of the form:

$$\psi_T = \sum_{i=1}^{n} \|y_i - t_i\|^2 \tag{2}$$

where $y_i$ is the network output. A long recognized bane of analysis of the error surface and the performance of training algorithms is the presence of multiple stationary points, including multiple minima. Empirical results with practical problems and training algorithms show that different initialization yields different networks [5][9]. Hence the issue of many minima is a real one. According to Auer *et al* [17], a single node network with *n* training pairs and $R^d$ inputs, could end up having $(\frac{n}{d})^d$ local minima. Hence, not only do multiple minima exist, but also, there may be huge numbers of them.

Different learning algorithms have staunch proponents who can always construct instances in which their algorithm performs better than most others. In practice, optimization algorithms that are used to minimize $\Psi_T(w)$ can be classified into four categories. The first three methods, gradient descent, conjugate gradients and quasi-Newton, are general optimization methods whose operation can be understood in the context of minimization of a quadratic error function [25][38[73]. Although the error surface is not quadratic, for differentiable node functions, it will be in a sufficiently small neighborhood of a local minimum. Such an analysis provides information about the behavior of the training algorithm over the span of a few iterations and also as it approaches its goal. The fourth method, that of Levenberg and Marquardt [31], is specifically adapted to minimization of an error function that arises from a squared error criterion of the form assumed. Backpropagation calculation of the gradient can be adapted easily to provide the information about the Jacobian matrix *J* needed for this method. A common feature of these training algorithms is the requirement of repeated efficient calculation of gradients [56].

Many of the conventional ANNs now being designed are statistically quite accurate but still leave a bad taste with users who expect computers to solve their problems accurately. The important drawback is that the designer has to specify the number of neurons, their distribution over several layers and the interconnection between them. Several methods have been proposed to automatically construct ANNs for reduction in network complexity that is to determine the appropriate number of hidden units, layers and learning rules [82]. Topological optimization algorithms such as Extentron [18], Upstart [41], Tiling

[70], Pruning [88] and Cascade Correlation [37] have their own limitations [5][104].

Evolutionary design of neural networks eliminates the tedious trial and error work of manually finding an optimal network [5][15][19][39][94-95][103]. The advantage of automatic design over manual design becomes clearer as the complexity of ANN increases. Evolutionary Artificial Neural Networks (EANN) provide a general framework for investigating various aspects of simulated evolution and learning. In EANN's, evolution can be introduced at various levels. At the lowest, it can be introduced into weight training, where ANN weights are evolved. At the next level, it can be introduced into neural network architecture adaptation, where the architecture (number of hidden layers, the number of hidden neurons and node transfer functions) is evolved. At the highest level, it can be introduced into the learning mechanism.

## 3.1 Meta Learning Evolutionary Artificial Neural Networks

One major problem with evolutionary algorithms is their inefficiency in fine tuning local search, although they are good at global searches [7]. The efficiency of evolutionary training can be improved significantly by incorporating a local search procedure into the evolution. Evolutionary algorithms are used first to locate a good region in the space and then a local search procedure is used to find a near optimal solution in this region. It is interesting to think of finding good initial weights as locating a good region in the space. Defining that the basin of attraction of a local minimum is composed of all the points, sets of weights in this case, which can converge to the local minimum through a local search algorithm, then a global minimum can easily be found by the local search algorithm if the evolutionary algorithm can locate any point, that is, a set of initial weights, in the basin of attraction of the global minimum. In Figure 5, $G_1$ and $G_2$ could be considered to be the initial weights as located by the evolutionary search, and $W_A$ and $W_B$ the corresponding final weights fine-tuned by the meta-learning technique.

Figure 6 illustrates the architecture of the Meta Learning Evolutionary Artificial Neural Network (MLEANN) and the general interaction mechanism with the learning mechanism evolving at the highest level on the slowest time scale [5]. All the randomly generated architectures of the initial population are trained by different learning algorithms (backpropagation - BP, scaled conjugate gradient - SCG, quasi-Newton algorithm - QNA and Levenberg Marquardt - LM) and evolved in a parallel environment. Parameters controlling the performance of the learning algorithm will be adapted (for example, the learning rate and the momentum for BP) according to the problem. The Architecture of the chromosome is presented in Figure 7. Figure 8 depicts the MLEANN algorithm.

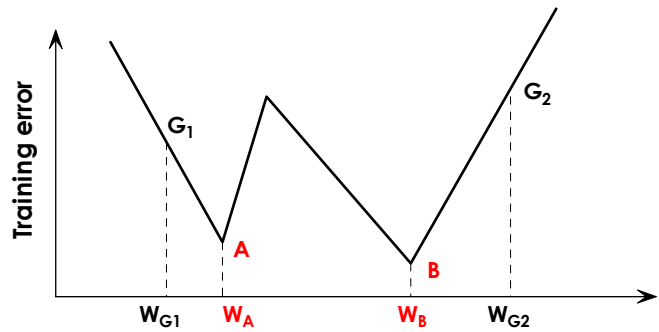

**Figure 5.** Fine tuning of weights using meta-learning

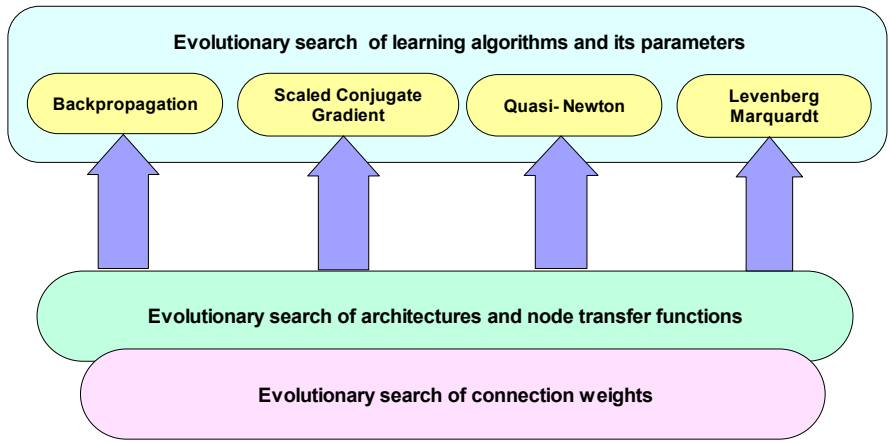

**Figure 6.** Interaction of various evolutionary search mechanisms

From the point of view of engineering, the decision about the level of evolution depends on what kind of prior knowledge is available. If there is more prior knowledge about EANN's architectures than that about their learning rules or a particular class of architectures is pursued, it is better to implement the evolution of architectures at the highest level because such knowledge can be used to reduce the search space and the lower levels of evolution of learning algorithms can be more biased towards this kind of architecture. On the other hand, the evolution of learning algorithms should be at the highest level if there is more prior knowledge available or a special interest in certain types of learning algorithm. Connection weights may be represented as binary strings represented by a certain length. The whole network is encoded by concatenation of all the connection weights of the network in the chromosome. A heuristic concerning the order of the concatenation is to put connection weights of the same node together.

Evolutionary architecture adaptation can be achieved by constructive [18][41] and destructive [88] algorithms. The former, which add complexity to the network starting from a very simple architecture until the entire network is able to learn the task. The latter start with large architectures and remove nodes and interconnections until the ANN is no longer able to perform its task. Then the last removal is undone. Direct encoding of the architecture makes the mapping simple but often suffers problems like scalability and implementation of crossover operators. For an optimal network, the required node transfer function (such as Gaussian, sigmoidal) could be formulated as a global search problem, which is evolved simultaneously with the search for architectures. For the neural network to be fully optimal, the learning algorithms have to be adapted dynamically according to the architecture and the given problem. Deciding the learning rate and momentum can be considered as the first attempt at adaptation of the local search technique (learning algorithm). The best learning algorithm will again be decided by the evolutionary search mechanism. Genotypes of the learning parameters of the different learning algorithms can be encoded as real-valued coefficients [15].

In Figure 7, for every learning algorithm parameter ($LR_2$), there is the evolution of architectures ($AR_1, AR_2.....AR_7....$) that proceeds on a faster time scale in an environment decided by the learning algorithm. For each architecture ($AR_3$), the evolution of connection weights ($WT_1, WT_2.....WT_5....$) proceeds at a faster time scale in an environment decided by the problem, the learning algorithm and the architecture.

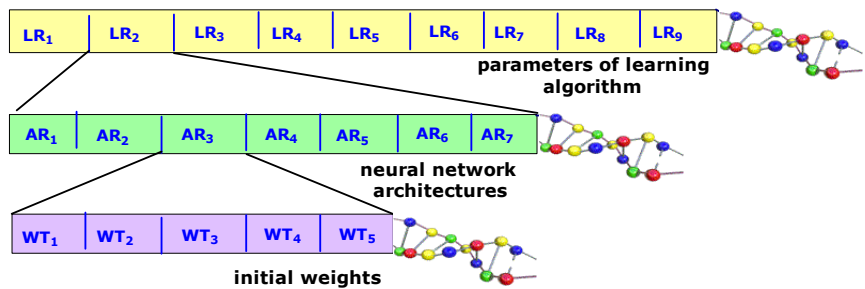

**Figure 7.** MLEANN chromosome architecture

The MLEANN approach has been applied for modelling three benchmark chaotic time series and the empirical results on test data sets clearly demonstrate the importance and efficacy of the meta learning approach for designing evolutionary neural networks [5][7]. Test results also demonstrate that MLEANN could outperform a Takagi-Sugeno [90] and Mamdani [68] fuzzy inference system which is learned using neural network learning methods.

> 1. *Set t=0 and randomly generate an initial population of neural networks with architectures, node transfer functions and connection weights assigned at random.*
> 2. *In a parallel mode, evaluate fitness of each ANN using BP/SCG/QNA and LM*
> 3. *Based on fitness value, select parents for reproduction*
> 4. *Apply mutation to the parents and produce offspring (s) for next generation. Refill the population back to the defined size.*
> 5. *Repeat step 2*
> 6. *STOP when the required solution is found or number of iterations has reached the required limit.*

**Figure 8**. MLEANN algorithm

The MLEANN approach was compared with the Cutting Angle Method (CAM) which is a deterministic global optimization technique [21]. This technique is based on theoretical results in abstract convexity [16]. It systematically explores the whole domain by calculating the values of the objective function $f(x)$ at certain points which are selected in such a way that the algorithm does not return to unpromising regions where function values are high. The new point is chosen where the objective function can potentially take the lowest value. The function is assumed to be Lipschitz, and the value of the potential minima is calculated based on both the distance to the neighbouring points and the function values at these points. This process can be seen as constructing the piecewise linear lower approximation of the objective function $f(x)$. With the addition of new points, the approximation $h_k(x)$ becomes closer to the objective function, and the global minimum of the approximating function $x^*$ converges to the global minimum of the objective function. The lower approximation, the auxiliary function $h_k(x)$, is called the saw-tooth cover of $f$. The MLEANN approach performed marginally better in terms of the lowest error on test sets. However CAM performed much faster when compared to the population-based MLEANN approach.

Selection of the architecture of a network (the number of layers, hidden neurons, activation functions and connection weights) and the correct learning algorithm is a tedious task for designing an optimal artificial neural network. Moreover, for critical applications and hardware implementations optimal design often becomes a necessity. Empirical results are promising and similar approach could be used for optimizing recurrent neural networks and other connectionist models. For the evolutionary search of architectures, it will be interesting to model as co-evolving [34] sub-networks instead of evolving the whole network. Further, it will be worthwhile to explore the whole population information of the final generation for deciding the best solution [103]. A fixed chromosome structure (direct encoding technique) was used to represent the connection weights, architecture, learning algorithms and its parameters. As the size of the network increases, the chromosome size grows. Moreover, implementation of crossover operator is often difficult due to production of non-functional offsprings. Parameterized encoding overcomes the problems with direct encoding but the search of architectures is restricted to layers. In the grammatical encoding a re-written grammar is encoded. The success will depend on the coding of grammar (rules). Cellular configuration might be helpful to explore the architecture of neural networks more efficiently. Gutierrez et al [45] has shown that their cellular automata technique performed better than direct coding.

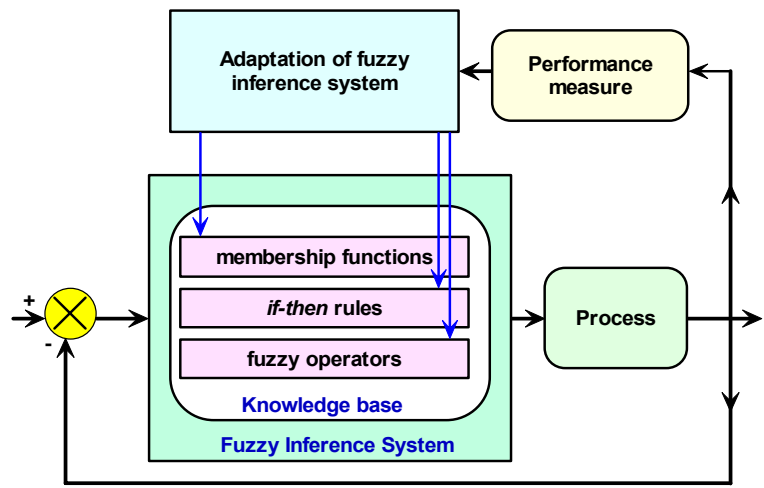

**Figure 9.** Architecture of adaptive fuzzy inference systems

## 4. Adaptation of Fuzzy Inference Systems

A conventional fuzzy controller makes use of a model of the expert who is in a position to specify the most important properties of the process. Expert knowledge is often the main source for designing Fuzzy Inference Systems (FIS) [81]. Figure 9 shows the architecture of the fuzzy inference system controlling a process. According to the performance measure of the problem environment, the membership functions, the knowledge base and the inference mechanism are to be adapted. Several research works continue to explore the adaptation of fuzzy inference systems [32][49][66-67][84][99]. These include the adaptation of membership functions, rule bases and the aggregation operators. They include but are not limited to:

- The self-organizing process controller by Procyk *et al* [83] which considered the issue of rule generation and adaptation.

- The gradient descent and its variants which have been applied to fine-tune the parameters of the input and output membership functions [100].
- Pruning the quantity and adapting the shape of input/output membership functions [101].
- Tools to identify the structure of fuzzy models [89].
- Fuzzy discretization and clustering techniques [105].
- In most cases the inference of the fuzzy rules is carried out using the '*min*' and '*max*' operators for fuzzy intersection and union. If the T-norm and T-conorm operators are parameterized then the gradient descent technique could be used in a supervised learning environment to fine-tune the fuzzy operators.

The antecedent of the fuzzy rule defines a local fuzzy region, while the consequent describes the behavior within the region via various constituents. The consequent constituent can be a membership function (Mamdani model) [68] or a linear equation (first order Takagi-Sugeno model) [90]. An easiest way to formulate the initial rule base is the grid partition method, as shown in Figure 10 where the input space is divided into multi-dimensional partitions and then assign actions to each of the partitions. The consequent parts of the rule represent the actions associated with each partition. It is evident that the MFs and the number of rules are tightly related to the partitioning and it encounters problems when we have a moderately large number of input variables (curse of dimensionality). Tree and scatter partition relieves the problem of exponential increase in the number of rules but orthogonality is often a major problem associated with these partitioning techniques [54].

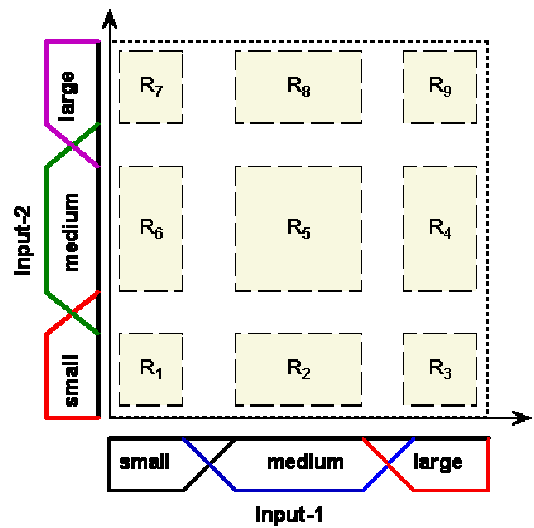

**Figure 10**. Grid partition: A simple *if-then* rule will appear as "*If input-1 is medium and input 2 is large then rule $R_8$ is fired*".

## 5. Evolutionary Fuzzy Systems

Adaptation of fuzzy inference systems using evolutionary computation techniques has been widely explored [11][32][76][79][85]. The evolutionary search of membership functions, rule base, fuzzy operators progress on different time scales to adapt the fuzzy inference system according to the problem environment. Figure 11 illustrates the general interaction mechanism with the evolutionary search of a fuzzy inference system (Mamdani, Takagi -Sugeno etc) evolving at the highest level on the slowest time scale. For each evolutionary search of fuzzy operators (for example, best combination of T-norm, T-conorm and defuzzification strategy), the search for the fuzzy rule base progresses at a faster time scale in an environment decided by the fuzzy inference system and the problem. In a similar manner, the evolutionary search of membership functions proceeds at a faster time scale (for every rule base) in the environment decided by the fuzzy inference system, fuzzy operators and the problem. The chromosome architecture is depicted in Figure 12.

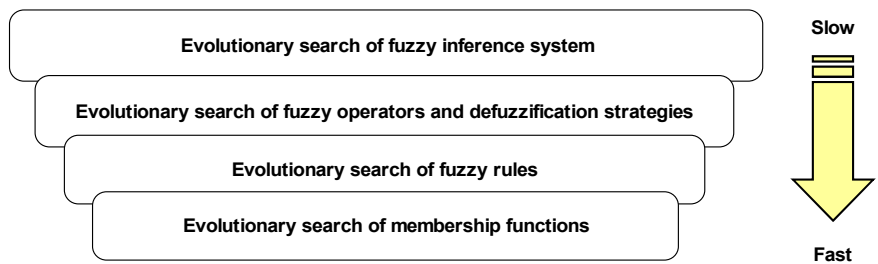

**Figure 11.** Interaction of the different evolutionary search mechanisms in the adaptation of fuzzy inference system

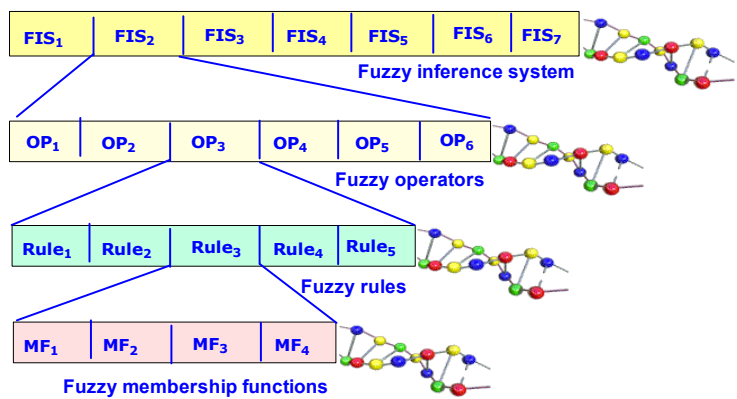

**Figure 12.** Chromosome representation of the adaptive fuzzy inference system

The automatic adaptation of membership functions is popularly known as self-tuning. The genome encodes parameters of trapezoidal, triangle, logistic, hyperbolic-tangent, Gaussian membership functions and so on [27].

The evolutionary search of fuzzy rules can be carried out using three approaches [32]. In the first (Michigan approach), the fuzzy knowledge base is adapted as a result of the antagonistic roles of competition and cooperation of fuzzy rules. Each genotype represents a single fuzzy rule and the entire population represents a solution. A classifier rule triggers whenever its condition part matches the current input, in which case, the proposed action is sent to the process to be controlled. The global search algorithm generates new classifier rules based on the rule strengths acquired during the entire process. The fuzzy behavior is created by an activation sequence of mutually-collaborating fuzzy rules. The entire knowledge base is built up by a cooperation of competing multiple fuzzy rules.

The second method (Pittsburgh approach) evolves a population of knowledge bases rather than individual fuzzy rules. Genetic operators serve to provide a new combination of rules and new rules. In some cases, variable length rule bases are used employing modified genetic operators for dealing with these variable length and position independent genomes. The disadvantage is the increased complexity of the search space and the additional computational burden, especially for online learning.

The third method (iterative rule learning approach) is similar to the first, with each chromosome representing a single rule, but contrary to the Michigan approach, only the best individual is considered to form part of the solution, the remaining chromosomes in the population are discarded. The evolutionary learning process builds up the complete rule base through an iterative learning process [44].

## 6. Cooperative Neuro-Fuzzy Systems

Hayashi et al [47] showed that a feedforward neural network could approximate any fuzzy-rule-based system and any feedforward neural network may be approximated by a rule-based fuzzy inference system [64]. A fusion of artificial neural networks and fuzzy inference systems has attracted growing interest amoung researchers in various scientific and engineering areas due to the growing need for adaptive intelligent systems to solve real world problems [2][4][6][8][10][33][43][46][52-54][59][62][66][78][98]. The advantages of a combination of neural networks and fuzzy inference systems are obvious [28-29][71]. An analysis reveals that the drawbacks pertaining to these approaches seem complementary and therefore, it is natural to consider building an integrated system combining the concepts. While the learning capability is an advantage from the viewpoint of a fuzzy inference system, the automatic formation of a linguistic rule base is an advantage from the viewpoint of neural networks. Neural network learning techniques could be used to learn the fuzzy inference system in a cooperative and an integrated environment. In this Section, three different types of cooperative neuro-fuzzy models are presented, namely fuzzy associative memories, fuzzy rule extraction using self-organizing maps and systems capable of learning fuzzy set parameters. Integrated neuro-fuzzy systems are presented in Section 7.

At the simplest level, a cooperative model can be thought of as a preprocessor wherein the ANN learning mechanism determines the fuzzy inference system membership functions or fuzzy rules from the training data. Once the FIS parameters are determined, ANN goes to the background.

Kosko's fuzzy associative memories [62], Pedryz's (*et al*) fuzzy rule extraction using self organizing maps [80] and Nomura's. (*et al*) systems capable of learning of fuzzy set parameters [75] are some good examples of cooperative neuro-fuzzy systems.

## 6.1 Fuzzy Associative memories

Kosko interprets a fuzzy rule as an association between antecedent and consequent parts [62]. If a fuzzy set is seen as a point in the unit hypercube and rules are associations, then it is possible to use neural associative memories to store fuzzy rules. A neural associative memory can be represented by its connection matrix. Associative recall is equivalent to multiplying a key factor with this matrix. The weights store the correlations between the features of the key, *k,* and the information part, *i*. Due to the restricted capacity of associative memories and because the combination of multiple connection matrices into a single matrix is not recommended due to severe loss of information, it is necessary to store each fuzzy rule in a single FAM. Rules with *n* conjunctively combined variables in their antecedents can be represented by *n* FAMs, where each stores a single rule. The FAMs are completed by aggregating all the individual outputs (maximum operator in the case of Mamdani fuzzy system) and a defuzzification component.

Learning can be incorporated in FAM as learning the weights associated with FAMs output or to create FAMs completely by learning. A neural network-learning algorithm determines the rule weights for the fuzzy rules. Such factors are often interpreted as the influence of a rule and are multiplied with the rule outputs. Rule weights can be replaced equivalently by modifying the membership functions. However, this could result in a misinterpretation of fuzzy sets and identical linguistic values might be represented differently in different rules. Kosko suggests a form of adaptive vector quantization technique to learn the FAMs. This approach is called differential competitive learning and is very similar to the learning in self-organizing maps.

Figure 13 depicts a cooperative neuro-fuzzy model where the neural network learning mechanism is used to determine the fuzzy rules, parameters of fuzzy sets, rule weights and so on. Kosko's adaptive FAM is a cooperative neuro-fuzzy model because it uses a learning technique to determine the rules and its weights. Its main disadvantage is the weighting of rules. Just because certain rules, do not have much influence does not mean that they are totally unimportant. Hence, the reliability of FAMs for certain applications is questionable. But because of their implementation simplicity, they are used in many applications.

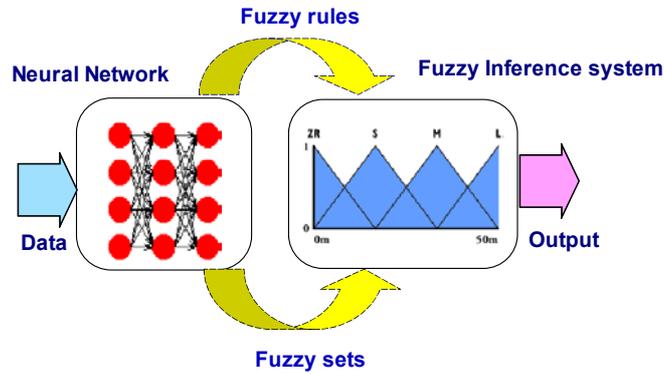

**Figure 13.** Cooperative neuro-fuzzy model

## 6.2 Fuzzy Rule Extraction Using Self Organizing Maps

Pedryz *et al* [80] used self-organizing maps with a planar competition layer to cluster training data, and they provide means to interpret the learning results. The learning results show whether two input vectors are similar to each other or belong to the same class. However, in the case of high-dimensional input vectors, the structure of the learning problem can rarely be detected in the two dimensional map. Pedryz et al provides a procedure for interpreting the learning results using linguistic variables.

After the learning process, the weight matrix *W* represents the weight of each feature of the input patterns to the output. Such a matrix defines a map for a single feature only. For each feature of the input patterns, fuzzy sets are specified by a linguistic description, *B* (one fuzzy set for each variable). They are applied to the weight matrix, *W,* to obtain a number of transformed matrices. Each combination of linguistic terms is a possible description of a pattern subset or cluster. To check a linguistic description, *B,* for validity, the transformed maps are intersected and a matrix *D* is obtained. Matrix *D* determines the compatibility of the learning result with the linguistic description *B*. $D^{(B)}$ is a fuzzy relation, and $d^{(B)}$ is interpreted as the degree of support of *B*. By describing $D^{(B)}$ by its *α*-cuts, $D_\alpha^B$ subsets of output nodes, whose degree of membership is at least *α,* so that the confidence of all patterns, $X_\alpha$, belong to the class described by *B* vanishes with decreasing *α*. Each B is a valid description of a cluster if $D^{(B)}$ has a non-empty *α*-cut $D_\alpha^B$. If the features are separated into input and output according to the application considered, then each B represents a linguistic rule, and by examining each combination of linguistic values, a complete fuzzy rule base can be created. This method also shows which patterns belong to a fuzzy rule, because they are not contained in any subset, $X_\alpha$. An important advantage compared to FAMs is that the rules are not weighted. The problem is with the determination of the number of output neurons and the *α* values for each learning problem. Compared to FAM, since the form of the membership function determines a crucial role in the performance, the data could be better exploited. Since Kosko's learning procedure does not take into account the neighborhood relation between the output neurons, perfect topological mapping from the input patterns to the output patterns might not be obtained. Thus the FAM learning procedure is more dependent on the sequence of the training data than the Pedryz *et al* procedure.

Pedryz *et al* initially determine the structure of the feature space and then the linguistic descriptions best matching the learning results, by using the available fuzzy partitions obtained. If a large number of patterns fit none of the descriptions, this may be due to an insufficient choice of membership functions and they can be determined anew. Hence, for learning the fuzzy rules, this approach is preferable compared to FAM [23]. Performance of this method still depends on the learning rate and the neighborhood size for weight modification, which is problem-dependant and could be determined heuristically. Fuzzy C-means algorithm also has been explored to determine the learning rate and neighborhood size [23][50].

## 6.3 Systems Capable of Learning Fuzzy Set Parameters

Nomura *et al* [75] proposed a supervised learning technique to fine-tune the fuzzy sets of an existing Sugeno type fuzzy system. Parameterized triangular membership functions were used for the antecedent part of the fuzzy rules. The learning algorithm is a gradient descent procedure that uses an error measure, *E,* (difference between the actual and target outputs) to fine-tune the parameters of the MF. Because the underlying fuzzy system uses neither a defuzzification procedure nor a non-differentiable t-norm to determine the fulfilment of rules, the calculation of the modifications of the MF parameters is trivial. The procedure is very similar to the delta rule for multilayer perceptrons. The learning takes place in an offline mode. For the input vector, the resulting error, *E,* is calculated and, based on that, the consequent parts (a real value) are updated. Then the same patterns are propagated again and only the parameters of the MFs are updated. This is done to take the changes in the consequents into account when the antecedents are modified. A severe drawback of this approach is that the representation of the linguistic values of the input variables depends on the rules they appear in. Initially, identical linguistic terms are represented by identical membership functions. During the learning process, they may be developed differently, so that identical linguistic terms are represented by different fuzzy sets. The proposed approach is applicable only to Sugeno type fuzzy inference system. Using a similar approach, Miyoshi et al [72] adapted fuzzy T-norm and T-conorm operators while Yager et al adapted the defuzzification operator using a supervised learning algorithm [102].

## 7. Integrated Neuro-Fuzzy Systems

In an integrated model, neural network learning algorithms are used to determine the parameters of fuzzy inference systems. Integrated neuro-fuzzy systems share data structures and knowledge representations. A fuzzy inference system can utilize human expertise by storing its essential components in a rule base and a database, and perform fuzzy reasoning to infer the overall output value. The derivation of *if-then* rules and corresponding membership functions depends heavily on the *a priori* knowledge about the system under consideration. However, there is no systematic way to transform the experiences of knowledge of human experts in to the knowledge base of a fuzzy inference system. There is also a need for the adaptability or some learning algorithms to produce outputs within the required error rate. On the other hand, the neural network learning mechanism does not rely on human expertise. Due to its homogenous structure, it is difficult to extract structured knowledge from either the weights or the configuration of the network. The weights of the neural network represent the coefficients of the hyper-plane that partition the input space into two regions with different output values. If this hyper-plane structure can be visualized from the training data the subsequent learning procedures in a neural network can be reduced. However, in reality, the *a priori* knowledge is usually obtained from human experts and it is most appropriate to express the knowledge as a set of fuzzy *if-then* rules and it is very difficult to encode into an neural network. Table 2 summarizes the comparison between neural networks and the fuzzy inference system [4][6]].

**Table 2**. Comparison between neural networks and fuzzy inference systems

| Artificial Neural Networks | Fuzzy Inference System |
| --- | --- |
| Prior rule-based knowledge cannot be used | Prior rule-base can be incorporated |
| Learning from scratch | Cannot learn (use linguistic knowledge) |
| Black box | Interpretable (if-then rules) |
| Complicated learning algorithms | Simple interpretation and implementation |

A common way to apply a learning algorithm to a fuzzy system is to represent it in a special neural network like architecture. Most of the integrated neuro-fuzzy models use a partitioning method (discussed in Section 4) to set up the initial rule base and then the learning algorithm is used to fine tune the parameters. However the conventional neural network learning algorithms (gradient descent) cannot be applied directly to such a system as the functions used in the inference process are usually non differentiable. This problem can be tackled by using differentiable functions in the inference system or by not using the standard neural learning algorithm. In Sections 7.1 and 7.2, how to model integrated neuro-fuzzy systems implementing Mamdani and Takagi - Sugeno FIS, is discussed.

## 7.1 Integrated Neuro-Fuzzy System (Mamdani FIS)

A Mamdani neuro-fuzzy system uses a supervised learning technique (backpropagation learning) to learn the parameters of the membership functions. The detailed function of each layer (as depicted in Figure 14) is as follows:

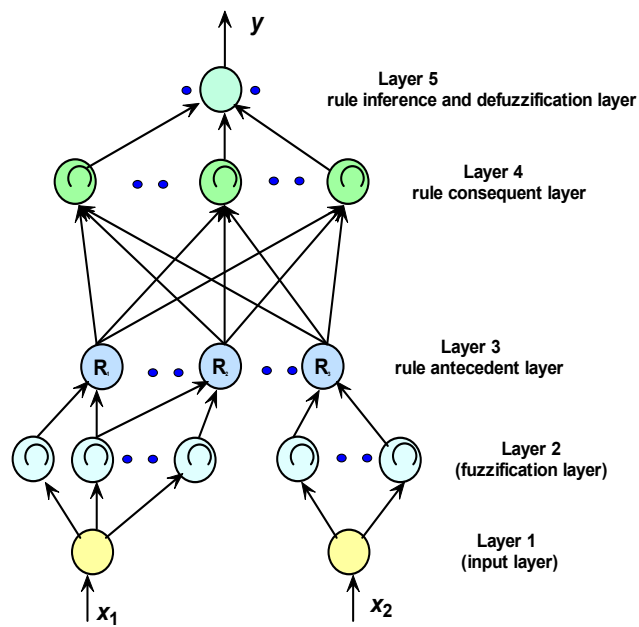

**Figure 14.** Mamdani neuro-fuzzy system

- Layer -1 *(input layer):* No computation is done in this layer. Each node, which corresponds to one input variable, only transmits input values to the next layer directly. The link weight in Layer 1 is unity.
- Layer-2 *(fuzzification layer):* Each node corresponds to one linguistic label (such as *excellent*, *good*) to one of the input variables in Layer 1. In other words, the output link represent the membership value, which specifies the degree to which an input value belongs to a fuzzy set, is calculated in layer 2. The final shapes of the MFs are fine tuned during network learning.
- Layer-3 *(rule antecedent layer):* A node represents the antecedent part of a rule. Usually a T-norm operator is used. The output of a Layer 3 node represents the firing strength of the corresponding fuzzy rule.

- Layer-4 *(rule consequent layer):* This node basically has two tasks: to combine the incoming rule antecedents and determine the degree to which they belong to the output linguistic label (for example, *high, medium, low*). The number of nodes in this layer are equal to the number of rules.

- Layer-5 *(Combination and defuzzification layer):* This node combines all the rules' consequents (normally using a T-conorm operator) and finally computes the crisp output after defuzzification.

## 7.2 Integrated Neuro-fuzzy system (Takagi-Sugeno FIS)

Takagi Sugeno neuro-fuzzy systems make use of a mixture of backpropagation to learn the membership functions and least mean square estimation to determine the coefficients of the linear combinations in the rule consequents. A step in the learning procedure has two parts: in the first, the input patterns are propagated, and the optimal conclusion parameters are estimated by an iterative least mean square procedure, while the antecedent parameters (membership functions) are assumed to be fixed for the current cycle through the training set; in the second, the patterns are propagated again, and in this epoch, backpropagation is used to modify the antecedent parameters, while the conclusion parameters remain fixed. This procedure is then iterated. The detailed functioning of each layer (as depicted in Figure 15) is as follows:

- Layers 1,2 and 3 functions the same way as Mamdani FIS.

- Layer 4 *(rule strength normalization):* Every node in this layer calculates the ratio of the *i*-th rule's firing strength to the sum of all rules' firing strength

$$\overline{w_i} = \frac{w_i}{w_1 + w_2}, i = 1,2..... . \tag{3}$$

- Layer-5 *(rule consequent layer):* Every node $i$ in this layer has a node function

$$\overline{w_i} f_i = \overline{w_i}(p_i x_1 + q_i x_2 + r_i), \tag{4}$$

where $\overline{w_i}$ is the output of layer 4, and $\{p_i, q_i, r_i\}$ is the parameter set. A well-established way is to determine the consequent parameters using the least means squares algorithm.

- Layer-6 *(rule inference layer)* The single node in this layer computes the overall output as the summation of all incoming signals:

$$Overall\ output = \sum_i \overline{w_i} f_i = \frac{\sum_i w_i f_i}{\sum_i w_i} \tag{5}$$

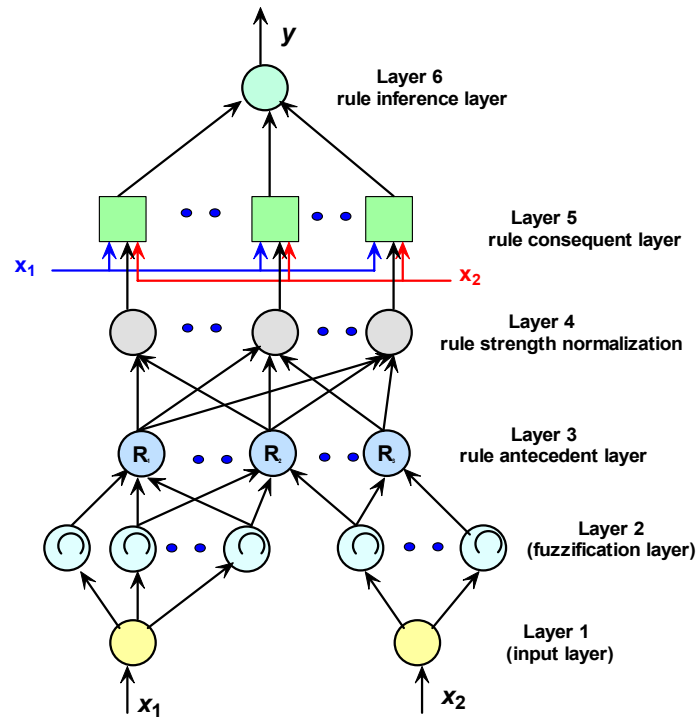

**Figure 15.** Takagi-Sugeno neuro-fuzzy system

Some of the integrated neuro-fuzzy systems are GARIC [22], FALCON [65], ANFIS [54], NEFCON, NEFCLASS, NEFPROX [74], FUN [91], SONFIN[55], FINEST[77][93], EFuNN [59-60] and EvoNF [1] [12]. A detailed review of the different integrated neuro-fuzzy models is presented in [6].

In ANFIS the adaptation (learning) process is only concerned with parameter level adaptation within fixed structures. For large-scale problems, it will be too complicated to determine the optimal premise-consequent structures, rule numbers etc. The structure of ANFIS ensures that each linguistic term is represented by only one fuzzy set. However the learning procedure of ANFIS does not provide the means to apply constraints that restrict the kind of modifications applied to the membership functions. When using Gaussian membership functions, operationally ANFIS can be compared with a radial basis function network.

NEFCON make use of a reinforcement type of learning algorithm for learning the rule base (structure learning) and a fuzzy backpropagation algorithm for learning the fuzzy sets (parameter learning). NEFCON system is capable of incorporating prior knowledge as well as learning from scratch. However the performance of the system will very much depend on heuristic factors like learning rate, error measure etc.

FINEST provides a mechanism based on the improved generalized modus ponens for fine tuning of fuzzy predicates and combination functions and tuning of an implication function. Parameterization of the inference procedure is very much essential for proper application of the tuning algorithm.

SONFIN is is adaptable to the users specification of required accuracy. Precondition parameters are tuned by backpropagation algorithm and consequent parameters by least mean squares or recursive least squares algorithms very similar to ANFIS.

EFuNN implements a Mamdani type of fuzzy rule base, based on a dynamic structure (creating and deleting strategy), and single rule inference, established on the winner-takes all rule for the rule node activation, with a one-pass training, instance based learning and reasoning. dmEFuNN is an improved version of the EFuNN capable of implementing Takagi-Sugeno fuzzy system, using several (m) of the highest activated rule nodes instead of one. The rule node aggregation is achieved by a C-means clustering algorithm.

FUN system is initialized by specifying a fixed number of rules and a fixed number of initial fuzzy sets for each variable and the network learns through a stochastic procedure that randomly changes parameters of membership functions and connections within the network structure Since no formal neural network learning technique is used it is questionable to call FUN a neuro-fuzzy system.

Sugeno-type fuzzy systems are high performers (less Root Mean Squared Error-RMSE) but often requires complicated learning procedures and are computationally expensive. However, Mamdani-type fuzzy systems can be modeled using faster heuristics but with a compromise on performance (high RMSE). There is always a compromise between performance and computational time. The data acquisition and preprocessing training data are also quite important for the success of neuro-fuzzy systems.

The success with integrating neural network and fuzzy logic and knowing their strengths and weaknesses, can be used to construct better neuro-fuzzy systems to mitigate the limitations and take advantage of the opportunities to produce more powerful hybrids than those that could be built with stand alone systems. As a guideline, for neuro-fuzzy systems to be at the top of the ladder, some of the major requirements are: fast learning (memory based - efficient storage and retrieval capacities), on-line adaptability (accommodating new features like inputs, outputs, nodes, connections), a global error rate and inexpensive computations (fast performance). As the problem become more complicated manual definition of neuro-fuzzy architecture/parameters becomes complicated. Especially for tasks requiring an optimal FIS, global optimization approach might be the best solution. In Section 8, EvoNF: a frame work for optimization of FIS using evolutionary algorithms and neural network learning technique is presented. EvoNF approach could be considered as a meta learning approach of evolutionary fuzzy systems.

## 8. Neuro-Fuzzy-Evolutionary (EvoNF) Systems

In an integrated neuro-fuzzy model, there is no guarantee that the neural network-learning algorithm will converge and the tuning of fuzzy inference system be successful. Optimization of fuzzy inference systems could be further improved using a meta-heuristic approach combining neural network learning algorithm and evolutionary algorithms. The proposed technique could be considered as a methodology to integrate neural networks, fuzzy inference systems and evolutionary search procedures [1] [3] [12].

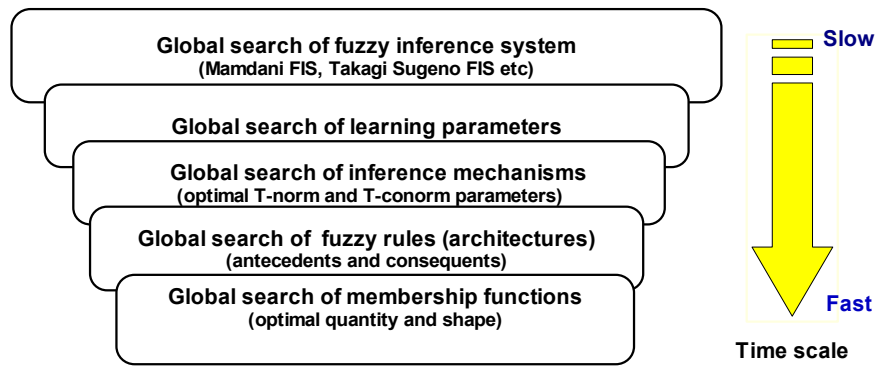

**Figure 16.** General computational framework for EvoNF

The EvoNF framework could adapt to Mamdani, Takagi-Sugeno or other fuzzy inference systems. The architecture and the evolving mechanism could be considered as a general framework for adaptive fuzzy systems, that is a fuzzy model that can change membership functions (quantity and shape), rule base (architecture), fuzzy operators and learning parameters according to different environments without human intervention. Solving multi-objective scientific and engineering problems is, generally, a very difficult goal. In these particular optimization problems, the objectives often conflict across a high-dimension problem space and may also require extensive computational resources. Proposed here is an evolutionary search procedure wherein the membership functions, rule base (architecture), fuzzy inference mechanism (T-norm and T-conorm operators), learning parameters and finally the type of inference system (Mamdani, Takagi-Sugeno etc.) are adapted according to the environment. Figure 15 illustrates the interaction of various evolutionary search procedures and shows that for every fuzzy inference system, there exists a global search of learning algorithm parameters, an inference mechanism, a rule base and membership functions in an environment decided by the problem. Thus, the evolution of the fuzzy inference system evolves at the slowest time scale while the evolution of the quantity and type of membership functions evolves at the fastest rate. The function of the other layers could be derived similarly.

The hierarchy of the different adaptation layers (procedures) relies on prior knowledge. For example, if there is more prior knowledge about the architecture than the inference mechanism then it is better to implement the architecture at a higher level. If a particular fuzzy inference system best suits the problem, the computational task could be reduced by minimizing the search space.

A typical chromosome of EvoNF would be as shown in Figure 17 and the detailed modelling process could be obtained from [1][12]. The chromosome architecture is very similar to to the chromosome structure mentioned in Figures 7 and 12.

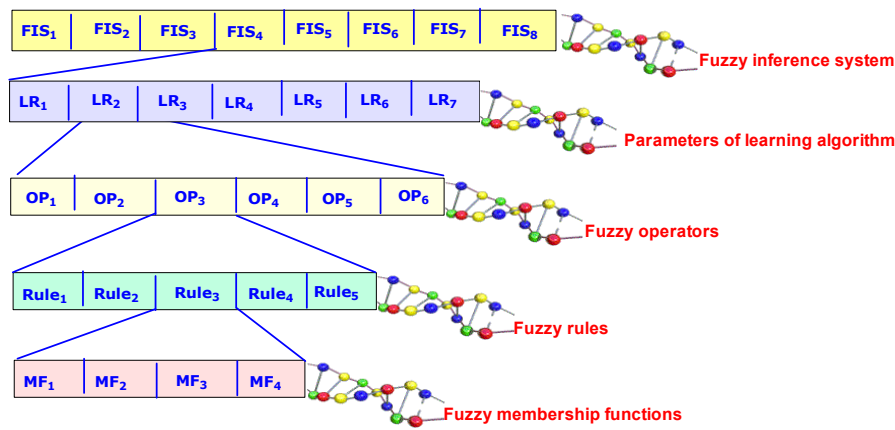

**Figure 17.** Chromosome structure of the EvoNF model

We have applied the proposed technique to the three well known chaotic time series. Fitness value is calculated based on the RMSE achieved on the test set. We have considered the best-evolved EvoNF model as the best individual of the last generation. We also explored different learning methods combining evolutionary learning and gradient descent techniques and the importance of tuning of different parameters. To reduce the computational complexity of the hierarchical search procedure, we reduced the search space by incorporating some priori knowledge. The genotypes were represented by real coding using floating-point numbers and the initial populations were randomly created. For all the three time series considered, EvoNF gave the best results on training and test sets [1] when compared to other integrated neuro-fuzzy models. Our experiments using the three different learning strategies also reveal the importance of fine-tuning the global search method using a local search method [3]. Figure 18 illustrates the comparison of EvoNF model with different integrated neuro-fuzzy models for predicting the Mackey Glass time series [1]. In Figure 18, test set RMSE values are given for each neuro-fuzzy model considered and an artificial neural network trained using BP.

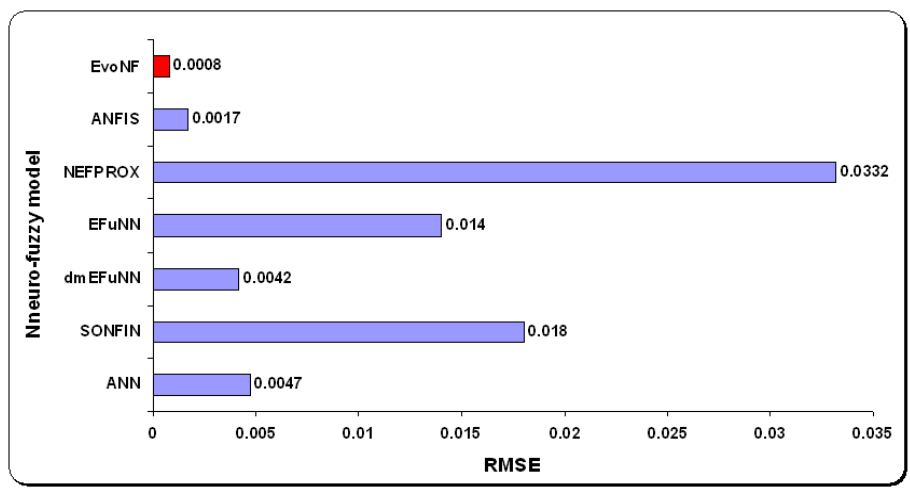

**Figure 18.** Comparison of EvoNF and some popular neuro-fuzzy models

## 9. Fuzzy Evolutionary Algorithms

Evolutionary algorithms are relatively easy to implement and, in general, their performance tends to be rather satisfactory in comparison with the small amount of knowledge about the problem they need in order to work. However, their success relies directly on the carefull selection of algorithm parameters, fitness function and so on. The use of fuzzy logic to translate and improve heuristic rules has also been applied to manage the resource of evolutionary algorithms such as population size and selection pressure as the algorithm greedily explores and exploits the search space [48]. The technique proposed by Lee [63] to perform a run-time tuning of population size and reproduction operators based on the fitness measures has shown large improvements in the computational run-time efficiency of the evolutionary search process. The fuzzy controller takes the inputs

$$\frac{average\ fitness}{best\ fitness}, \frac{worst\ fitness}{average\ fitness}, \Delta best\ fitness$$

and gives Δpopulation size, Δcrossover rate and Δmutation rate to control the evolutionary algorithm parameters. The ranges of the parameter changes are also limited to remain within certain bandwidths. This technique could improve not only the search efficiency and convergence but also sometimes could avoid premature convergence due to lack of diversity in the population.

As mentioned in Section 5, the two ingredients of soft computing, evolutionary computation and fuzzy inference systems, could be integrated in a way that makes them benefit from one another.

## 10. Soft Computing and Probabilistic Reasoning

A common feature of soft computing technology and the probabilistic reasoning system is their depature from classical reasoning and modeling approaches which are highly based on analytical models, crisp logic and deterministic search. In the probabilistic modeling process, risk means the uncertainty for which the probability distribution is known. The probabilistic models are used for protection against adverse uncertainty and exploitation of propitious uncertainty.

In a probabilistic neural network (Bayesian learning) probability is used to represent uncertainty about the relationship being learned. Before any data is seen the *prior* opinions about what the true relationship might be can be expressed in a probability distribution over the network weights that define this relationship. After a look at the data, revised opinions are captured by a *posterior* distribution over network weights. Network weights that seemed plausible before, but which do not match the data very well, are now seen as being much less likely, while the probability for values of the weights that do fit the data well have increased. Typically, the purpose of training is to make predictions for future cases in which only the inputs to the network are known. The result of conventional network training is a single set of weights that can be used to make such predictions.

Several research work has exposed the complementary features of probabilistic reasoning and fuzzy theory [26]. The development of the theory of belief of a fuzzy event by Smets [87] helped to establish the orthogonality and complementarity between probabilistic and possibilistic methods.

## 11. Conclusions

It is predicted that, in the 21$^{st}$ century, the fundamental source of wealth will be knowledge and communication rather than natural resources and physical labour. With the exponential growth of information and complexity in this world, intelligent systems are needed that could learn from data in a continuous, incremental way, and grow as they operate, update their knowledge and refine the model through interaction with the environment. The intelligence of such systems could be further improved if the adaptation process could learn from successes and mistakes and that knowledge be applied to new problems.

This chapter has presented some of the architectures and perspectives of hybrid intelligent systems involving neural networks, fuzzy inference systems, evolutionary computation and probabilistic reasoning. The hybrid soft computing approach has many important practical applications in science, technology, business and commercial. Compared to the individual constituents (NN, FIS, EC ans PR) hybrid soft computing frameworks are relatively young. As the strengths and weakness of different hybrid architectures are understood, it will be possible to use them more efficiently to solve real world problems.

The integration of different intelligent technologies is the most exciting fruit of modern artificial intelligence and is an active area of research. While James Bezdek [24] defines intelligent systems in a frame called computational intelligence, Lotfi Zadeh [108] explains the same by using the soft computing framework. Integration issues range from different techniques and theories of computation to problems of exactly how best to implement hybrid systems. Like most biological systems which can adapt to any environment, adaptable intelligent systems are required to tackle future complex problems involving huge data volume. Most of the existing hybrid soft computing frameworks rely on several user specified network parameters. For the system to be fully adaptable, performance should not be heavily dependant on user-specified parameters.

For optimizing neural networks and fuzzy inference systems, there is perhaps no better algorithm than evolutionary algorithms. However, the real success in modeling such systems will directly depend on the genotype representation of the different layers. The population-based collective learning process, self-adaptation, and robustness are some of the key features of evolutionary algorithms when compared to other global optimization techniques. Evolutionary algorithms attract considerable computational effort especially for problems involving complexity and huge data volume. Fortunately, evolutionary algorithms work with a population of independent solutions, which makes it easy to distribute the computational load among several processors.

## Acknowledgements

Author is grateful to Professor Lakhmi Jain (University of South Australia, Adelaide) and the three referees for the technical comments, which improved the clarity of this chapter.